\documentclass{article} \usepackage{iclr2024_conference,times}

\usepackage{amsmath,amsfonts,bm}

\def\eqref#1{equation~\ref{#1}}

\def\1{\bm{1}}

\DeclareMathAlphabet{\mathsfit}{\encodingdefault}{\sfdefault}{m}{sl}
\SetMathAlphabet{\mathsfit}{bold}{\encodingdefault}{\sfdefault}{bx}{n}

\newcommand{\softmax}{\mathrm{softmax}}

\usepackage{hyperref}
\usepackage{cleveref}
\usepackage{caption, subcaption}
\usepackage{url}
\usepackage{cases}
\usepackage{graphicx}
\usepackage{amsmath, amssymb}
\usepackage{bbm}
\usepackage{rotating}
\usepackage{multirow}
\usepackage{array}
\usepackage{glossaries}
\usepackage{tabularray}
\usepackage{colortbl}
\usepackage{makecell}
\usepackage{wrapfig}
\usepackage{mathtools}
\UseTblrLibrary{booktabs}
\usepackage{pifont}\newcommand{\cmark}{\ding{51}}\newcommand{\xmark}{\ding{55}}

\title{Multi-Modal Contrastive Learning \\ for Online Clinical Time-Series Applications}

\author{Fabian Baldenweg, Manuel Burger, Gunnar Rätsch, Rita Kuznetsova\\
Department of Computer Science, ETH Zürich, Switzerland \\
\texttt{\{bafabian,burgerm,raetsch,mkuznetsova\}@ethz.ch} \\
}

\newacronym[plural=ICUs, firstplural={Intensive Care Units (ICUs)}]{icu}{ICU}{Intensive Care Unit}
\newacronym{ncl}{NCL}{Neighborhood Contrastive Learning}
\newacronym{mlp}{MLP}{Multi-Layer Perceptron}
\newacronym{gru}{GRU}{Gated Recurrent Unit}

\newcommand{\N}{\mathbb{N}}
\newcommand{\ReSet}{\mathbb{R}}

\newcommand{\tm}{\textit{time}}
\newcommand{\gru}{\text{GRU}}
\newcommand{\lm}{\text{LM}}
\newcommand{\mlp}{\text{MLP}}
\newcommand{\norm}[1]{\left\lVert #1 \right\rVert}

\iclrfinalcopy \begin{document}

\setlength{\parskip}{1mm}

\maketitle

\begin{abstract}
Electronic Health Record (EHR) datasets from Intensive Care Units (ICU) contain a diverse set of data modalities. While prior works have successfully leveraged multiple modalities in supervised settings, we apply advanced self-supervised multi-modal contrastive learning techniques to ICU data, specifically focusing on clinical notes and time-series for clinically relevant online prediction tasks. We introduce a loss function \emph{Multi-Modal Neighborhood Contrastive Loss (MM-NCL)}, a \emph{soft} neighborhood function, and showcase the excellent linear probe and zero-shot performance of our approach.
\end{abstract}

\section{Introduction}

Electronic Health Record (EHR) data from Intensive Care Units (ICUs) has emerged as a valuable resource for predicting clinically relevant quantities in recent years~\citep{circews-hyland-nature20, respews-hueser-medrxiv24, yeche2022hiridicubenchmark, pace2022poetree}. However, the diverse nature of EHR data, encompassing different modalities such as clinical notes and time series, presents a challenge for effective utilization. The majority of models leveraging multiple modalities rely on supervised learning~\cite{husmann2022importance, jain2023knowledge, khadanga2019using}, necessitating separate training for each task, demanding substantial amounts of annotated data, and learning only task-specific modality interactions. There remains a gap for an architecture, which fuses modalities in a task-agnostic manner.

To address these challenges, there is growing interest in developing self-supervised approaches~\citep{oord2019representation, yeche2021neighborhood} that can learn task-agnostic representations, thereby reducing or eliminating the dependency on annotated data. Encouragingly, contrastive learning has proven successful in creating such multi-modal representations for text and images without task-specific training~\citep{radford2021learning, wang2022medclip, li2023blip2}. \citet{radford2021learning} even demonstrate strong zero-shot classification performance based on their multi-modal shared latent space.

\textbf{Our Contribution}: Motivated by initial exploration~\citep{radford2021learning, king2023multimodal}, we aim to apply multi-modal contrastive learning, specifically focusing on clinical notes and medical time-series, while aiming to improve performance for online prediction tasks~\citep{yeche2021neighborhood}. We introduce a loss function titled \emph{Multi-Modal Neighborhood Contrastive Learning (MM-NCL)} together with a novel \emph{soft} neighborhood function.
We showcase the strong linear probe and zero-shot performance of our approach on in-hospital mortality and, most importantly, decompensation tasks. To the best of our knowledge, our decompensation results represent the best successful benchmarked zero-shot performance on an online ICU prediction task.

\section{Related Work}

\paragraph{Learning on Medical Time-Series}

A wide range of research has been conducted on supervised learning applied to ICU time series~\citep{Harutyunyan2019, pmlr-v225-kuznetsova23a, yeche2022hiridicubenchmark, vandewaterYetAnotherICUBenchmark2023}.
Uni-modal contrastive methods~\citep{yeche2021neighborhood, zhang2022selfsupervised, pmlr-v174-weatherhead22a}, on the other hand, are more closely related to this work. 
They often rely on \emph{InfoNCE}~\citep{oord2019representation} or a variant thereof to pull augmented views of the same data closer together and push different samples apart in the representation space~\cite{Liu2023Survey}.

\paragraph{Multi-modal Learning}

Multi-modal contrastive learning was popularized by CLIP~\citep{radford2021learning} contrasting images and captions, leveraging it to train models for zero-shot classification, and laying the groundwork for state-of-the-art image generators~\citep{rombach2022highresolution}.
Adaptations of CLIP have been proposed for video input~\citep{liu2023revisiting}, multi-lingual text~\citep{chen2022altclip}, and pre-trained uni-modal encoders~\citep{li2023blip2}.
In the medical domain, MedCLIP~\citep{wang2022medclip} adopts this framework and applies it to X-ray images and clinical notes. To overcome the scarcity of paired multi-modal samples they propose a soft assignment based on a text similarity scoring.
\citet{Li2022} used multi-modal contrastive learning to align medical time-series and hyperbolic embeddings of ICD codes, while MedFuse~\citep{hayat2023medfuse} does so with images.
\citet{king2023multimodal} recently explored multi-modal contrastive learning between clinical notes and medical time-series, however, their approach differs from ours in several aspects, their architecture is only suitable for offline tasks, and we show vastly better performance especially in the zero-shot setting.

\section{Methods}
\paragraph{Notation} We consider a patient dataset of multiple paired time series $X^S$ and clinical note series $X^T$, where each pair represents a patient's stay in the ICU.
The time series contains $d_v$ hourly vital signs of patients in the ICU, while the clinical note series is sparse across time.
$X^S_t \in \ReSet^{d_v}$ is step $t$ of $X^S$ and $X^S[t_1:t_2]$ for $t_1, t_2 \in \N, t_1 \leq t_2$ denotes the sub-sequence of $X^S$ between steps $t_1$ (inclusive) and $t_2$ (exclusive).
Let $X^T_{i,j}$ denote the $j$-th note in ICU stay $i$.
A batch consists of $K$ pairs of texts $X^T_{i,j}$ and time series $X^S_i$ $, i \in \{1, 2, \ldots K-1\}$.
A sub-sequence of length $w$ (window size) of each time-series near the creation time of the note is fed to the time-series encoder.
Let $\tm(X^T_{i,j})$ be the creation time of note $X^T_{i,j}$. Then, for each note $X^T_{i,j}$, a target time $\tau_i$ is drawn uniformly at random from $[\tm(X^T_{i,j}) - a, \tm(X^T_{i,j}) + b]$ where $a$ and $b$ may depend on the type of notes.
The final time-series $\tilde{X}^S_{i,j}, i \in \{1, \ldots, K-1\}$ where $\tilde{X}^S_{i,j,\tau} = X^S_i[\tau_i - w:\tau_i]$ are fed to the model.

\begin{figure}[t]
    \centering
    \includegraphics[width=\textwidth]{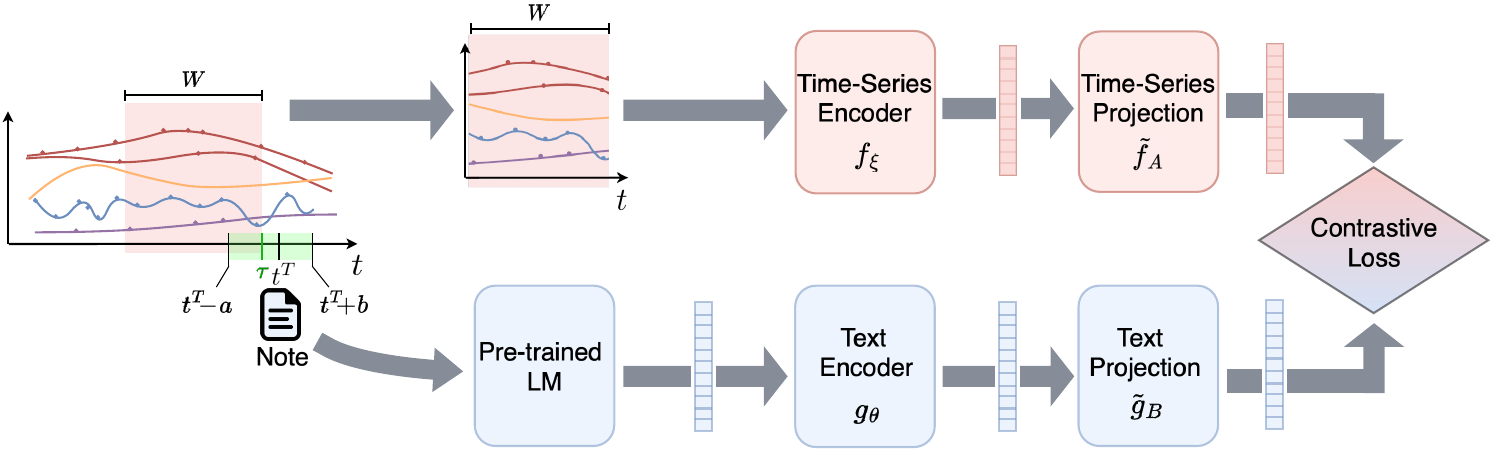}
    \caption{Training pipeline}
    \label{fig:architecture}
\end{figure}

\paragraph{Model Architecture} Our model (see \Cref{fig:architecture}) consists of a time-series encoder $f_\xi(X^S) = \gru_\xi(X^S)$, a time-series projection $\tilde{f}_A(x) = W_A x$,
a text encoder $g_\theta(X^T) = \textit{concat}(\mlp_\theta(\lm(X^T)), \lm(X^T))$ and a text projection $\tilde{g}_B(x) = W_B x$. $\gru_\xi$ is the last hidden state of a Gated Recurrent Unit~\citep{cho2014learning} with parameters $\xi$. $\mlp_\theta$ is a multi-layer perceptron (MLP) with parameters $\theta$ and $\lm$ is a pre-trained language model~\citep{huang2020clinicalbert}, which we use to compute a single representation vector per clinical note. $\textit{concat}$ concatenates two vectors. $W_A, W_B$ are trainable matrices.

\paragraph{Loss}
Based on \citet{yeche2021neighborhood} we propose a \emph{Multi-Modal Neighborhood Contrastive Loss (MM-NCL)} $\mathcal{L}_{MM-NCL}$.
We also do experiments with our pipeline using the original loss from CLIP~\citep{radford2021learning} (and refer to it using \emph{MM-InfoNCE}).

Let $B$ be the set of index tuples $\iota = (i, j, \tau)$ in a batch of size $K$, containing stay index, note index, and target time. Let $\nu > 0$ be a trainable temperature parameter and the following define the normalized embeddings ($norm(x) := x / \norm{x}$) of the time-series $\mathbf{h}_\iota^S$ and the texts $\mathbf{h}_\iota^T$ passed to the contrastive objective:

\begin{equation}
    \mathbf{h}_\iota^S \coloneqq norm\bigl( \tilde f_A(f_\xi(X^S_{i, j, \tau})) \bigr) \in \ReSet^c
    , \;\;\;
    \mathbf{h}_\iota^T \coloneqq norm\bigl( \tilde g_B(g_\theta(X^T_{i,j,\tau})) \bigr) \in \ReSet^c \label{eqn:embeddings}
\end{equation}

\emph{MM-NCL} consists of two components: (\textbf{i}) The neighborhood aware loss $\mathcal L_A$ (Eqn.~\ref{eqn:loss-la}) and (\textbf{ii}) the neighborhood discriminative loss $\mathcal L_D$ (Eqn.~\ref{eqn:loss-ld}). Eqn.~\ref{eqn:neighborhood-1} and~\ref{eqn:neighborhood-2} define a novel \emph{soft} neighborhood function relating neighboring clinical notes and time-series windows w.r.t. their distance in time:

\begin{align}
    \mathcal{L}_A &\coloneqq \sum_{l \in B} \sum_{m  \in B} - \frac{N_{l,m}}{2K} \left(\log{\frac{\exp{(\mathbf{h}_{l}^S \cdot \mathbf{h}_{m}^T / \nu)}}{\sum_{n \neq l}^B \exp{(\mathbf{h}_{l}^S \cdot \mathbf{h}_{n}^T / \nu)}}} + \log{\frac{\exp{(\mathbf{h}_{l}^T \cdot \mathbf{h}_{m}^S / \nu)}}{\sum_{n \neq l}^B \exp{(\mathbf{h}_{l}^T \cdot \mathbf{h}_{n}^S / \nu)}}} \right) \label{eqn:loss-la} \\
    \mathcal{L}_D &\coloneqq \sum_{l \in B} -\frac{1}{2K} \left( \log \frac{\exp(\mathbf{h}^S_{l} \cdot \mathbf{h}^T_{l} / \nu)}{\sum_{m}^B \mathbbm{1}^N_{l,m} \exp(\mathbf{h}_{l}^S \cdot \mathbf{h}_{m}^T / \nu)} + \log \frac{\exp(\mathbf{h}^T_{l} \cdot \mathbf{h}^S_{l} / \nu)}{\sum_{m}^B \mathbbm{1}^N_{l,m} \exp(\mathbf{h}_{l}^T \cdot \mathbf{h}_{m}^S / \nu)} \right) \label{eqn:loss-ld}\\
    \text{where}& \nonumber \\
    N_{l,m} &\coloneqq \frac{\tilde{N}(l, m)}{\sum_{n \in B} \tilde N(l,m)}, \hspace{2cm}\mathbbm{1}^N_{l,m} := \begin{cases}
    	1, & \text{if } N_{l,m} \neq 0 \\
    	0, & \text{if } N_{l,m} = 0
    \end{cases} \label{eqn:neighborhood-1} \\
\tilde N(l, m) &\coloneqq \begin{cases}
    \frac{\beta}{\beta + |\tau_m - \tau_l|}, & \text{if } i_l = i_m \land |j_l - j_m| \leq 1 \\
    0, & \text{otherwise}
    \end{cases} \label{eqn:neighborhood-2}
\end{align}

$\beta \in \ReSet^{\geq 1}$ is a hyperparameter (defining the soft neighborhood decay w.r.t. temporal distance) and the final loss function is a linear combination with trade-off hyperparameter $\alpha$:

\begin{equation}
    \mathcal{L}_{MM-NCL} \coloneqq \alpha \mathcal{L}_A + (1 - \alpha) \mathcal{L}_D, \hspace{1cm} \alpha \in (0, 1]
\end{equation}

While~\citet{yeche2021neighborhood} have proposed the separation of a contrastive loss into neighborhood \emph{aware} and \emph{distriminative} components, we have expanded their definition to the multi-modal setting and introduce a soft neighborhood function, where they have only considered the binary case.

\vspace{-0.4cm}
\section{Results and Discussion}
\label{sec:results}

\paragraph{Experimental Setup}
We use the time-series features, cohort selection, splits, and label definitions from MIMIC-III Benchmark~\citep{Harutyunyan2019}
and extracted the clinical notes from the MIMIC-III dataset~\citep{Johnson2016}.
We benchmark in-hospital mortality and decompensation as defined by~\citet{Harutyunyan2019}.
In-hospital mortality is an offline binary classification of predicting patient mortality after the first 48 hours of stay in the ICU.
Decompensation is an hourly \emph{online} binary classification task to predict the onset of death in the next 24 hours.
More details in Appendix~\ref{app:data} and~\ref{app:training}.

\paragraph{Evaluation}
Note that for all results, while the contrastive pretraining considers multiple modalities, for inference only time-series data is passed to the model. This is different from some supervised baselines~\citep{husmann2022importance, khadanga2019using} and is more suitable for an online deployment scenario, where measurements and lab results are naturally stored in databases, but clinical notes require a physician to analyse the respective data streams and write the note, before they would become visible to the system.
We evaluate linear probes~\citep{alain2017understanding} on the output of the frozen base time-series encoder $f_\xi(X^S)$. 
Further, we consider zero-shot classification by scoring the alignment of an embedded time-series window with class-specific text prompts~\citep{radford2021learning}. Time-series windows are classified by their similarity with positive (e.g. \textit{"patient died"}) and negative prompt ensembles (e.g. \textit{"patient survived"}, more examples in App.~\ref{app:model-prompts}).
Let $\mathcal{P}^T_+$ and $\mathcal{P}^T_-$ be the set of positive and negative text prompts.
Then the zero-shot probabilities $\hat{y}_{zs}$ for a time-series window $X^S$, encoded to $\mathbf{h^S}$ as in Eqn.~\ref{eqn:embeddings}, are (Eqn.~\ref{eqn:zs-classification}):

\begin{equation}
    \label{eqn:zs-classification}
    \hat{y}_{zs} = \softmax(\mathbf{h^S} \cdot \mathbf{p}^T_+,\; \mathbf{h^S} \cdot \mathbf{p}^T_-) \;\;\;\text{where}\;\;\; \mathbf{p}^T_{+/-} \coloneqq \frac{1}{|\mathcal{P}^T_{+/-}|} \sum_{p \in \mathcal{P}^T_{+/-}}norm{\bigl( g_B(g_\theta(p)) \bigr)}
\end{equation}

\begin{table*}[tb]
\small
    \centering
    \caption{Model Performance Comparison. All values are in \% and denoted as $mean \pm std$. Bold marks the best result in each section. \textit{MM-Train.} and \textit{-Infer.} mark if multiple modalities are used for training and inference. Missing values are not provided by the respective references.}
    \vspace{0.1cm}
    \begin{tabular}{l c c l l l l}
    \toprule
    \bf Method & \textit{MM} & \textit{MM} &  \multicolumn{2}{c}{\textbf{Mortality}} & \multicolumn{2}{c}{\textbf{Decompensation}}\\
    \cmidrule(lr){4-5} \cmidrule(lr){6-7}
      & \textit{Train.} & \textit{Infer.} & \textit{AuPRC} & \textit{AuROC} & \textit{AuPRC} & \textit{AuROC} \\

\midrule
    \multicolumn{5}{l}{\textbf{Supervised}} \\
    \arrayrulecolor{lightgray}\midrule\arrayrulecolor{black}
    \cite{Harutyunyan2019} & \xmark & \xmark & 50.1 $\pm$ 1.3 & 86.1 $\pm$ 0.3 & 34.1 $\pm$ 0.5 & 90.7$\pm$0.2 \\
    \cite{khadanga2019using} & \cmark & \cmark & 52.5$\pm$1.3 & 86.5$\pm$0.4 & 34.5$\pm$0.7 & 90.7$\pm$0.7 \\
    \cite{husmann2022importance} & \cmark & \cmark & \textbf{52.7$\pm$1.0} & \textbf{87.1$\pm$0.6} & \textbf{39.7$\pm$0.6} & \textbf{92.2$\pm$0.2} \\

\midrule
    \multicolumn{5}{l}{\textbf{Self-Supervised Linear Probes}} \\
    \arrayrulecolor{lightgray}\midrule\arrayrulecolor{black}
    \cite{yeche2021neighborhood} & \xmark & \xmark & - & - & 31.2 $\pm$ 0.5 & 88.9 $\pm$ 0.3 \\
\cite{king2023multimodal} & \cmark & \xmark & 40.2 $\pm$ 5.3 &  82.8 $\pm$ 2.0 & - & - \\

    MM-NCL (ours) & \cmark & \xmark & \textbf{52.1 $\pm$ 0.5} & \textbf{85.9 $\pm$ 0.2} & \textbf{32.6 $\pm$ 1.2} & \textbf{90.2 $\pm$ 0.4} \\ 

\midrule
    \multicolumn{5}{l}{\textbf{Self-Supervised Zero-Shot}} \\
    \arrayrulecolor{lightgray}\midrule\arrayrulecolor{black}
    \cite{king2023multimodal} & \cmark & \xmark & 21.4 $\pm$ nan &  70.9 $\pm$ nan & - & - \\
    MM-InfoNCE (ours) & \cmark & \xmark & \textbf{48.3 $\pm$ 1.4} & \textbf{83.2 $\pm$ 0.63} & 26.9 $\pm$ 2.2 & \textbf{87.8 $\pm$ 0.2} \\
MM-NCL (ours) & \cmark & \xmark & 45.1 $\pm$ 2.8 & 80.0 $\pm$ 2.4 & \textbf{30.9 $\pm$ 0.7} & 87.4 $\pm$ 0.7 \\ \bottomrule
    \end{tabular}
    \label{tab:performance-comparison}
\end{table*}

\paragraph{Results} \Cref{tab:performance-comparison} shows comparisons to a supervised time-series baseline~\citep{Harutyunyan2019} and supervised multi-modal baselines~\citep{khadanga2019using, husmann2022importance}. Further, we compare to self-supervised results for online predictions~\citep{yeche2021neighborhood} and self-supervised multi-modal results~\citep{king2023multimodal}. We strongly outperform prior work by~\citet{king2023multimodal} in both the probed and the zero-shot setting. Our probed results on mortality can even compete with a strong supervised multi-modal baseline by~\citet{husmann2022importance}, while on decompensation we can slightly improve upon the results by~\citet{yeche2021neighborhood}.

In the zero-shot setting, we present the first results on an online patient prediction task (decompensation). Our loss function for multi-modal neighborhood contrastive learning in online settings achieves a zero-shot performance getting close to probed results. Additionally, we vastly outperform the only available prior result on multi-modal contrastive learning for time-series and clinical notes on mortality by~\citet{king2023multimodal}. The difference in performance on mortality for MM-InfoNCE and MM-NCL can be attributed to the nature of our proposed loss function focusing on more local, clinically relevant~\citep{yeche2021neighborhood}, online patient state changes, while on the offline mortality prediction global alignment is favored.

\begin{wrapfigure}[18]{r}{0.45\textwidth}
    \vspace{-1cm}
    \centering
    \includegraphics[width=0.45\textwidth]{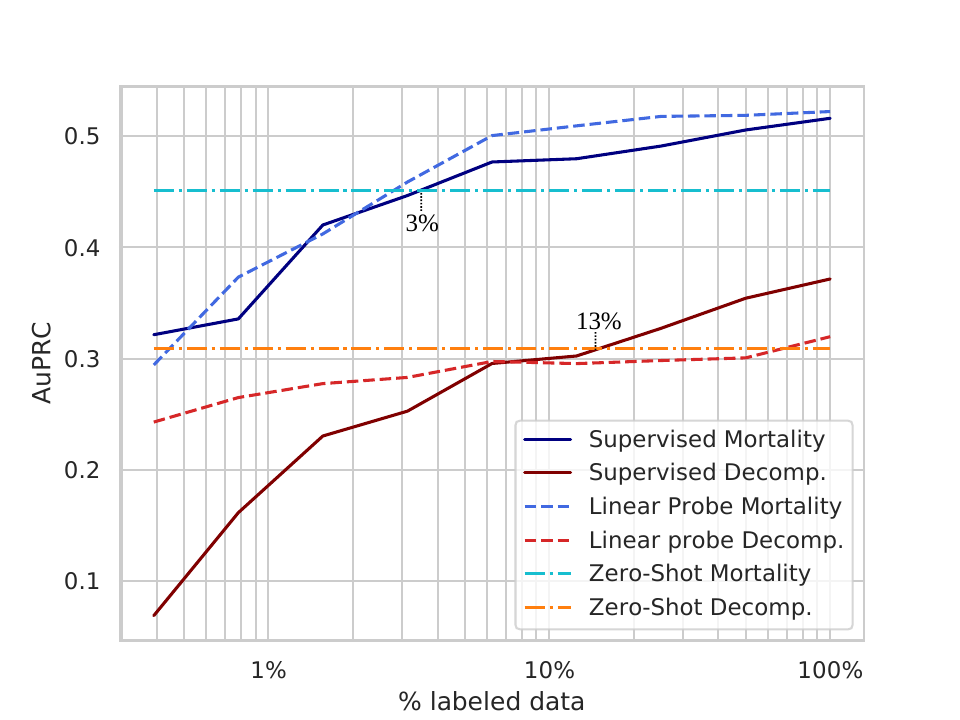}
    \caption{AuPRC when training with reduced labels, x-axis shows the percentage of of labels used from the full training set. All results were obtained using the same time-series architecture. We mark the percentage, where supervised outperforms zero-shot.}
    \label{fig:reduced_labels_performance}
\end{wrapfigure}

\paragraph{Scarce label setting} \Cref{fig:reduced_labels_performance} compares supervised, linear probe and zero-shot results on training sets with reduced labels. It clearly shows the superiority of zero-shot predictions in the scarce label regime.

\begin{figure}
    \centering
\begin{subfigure}[b]{0.49\textwidth}
        \centering
        \includegraphics[width=\textwidth]{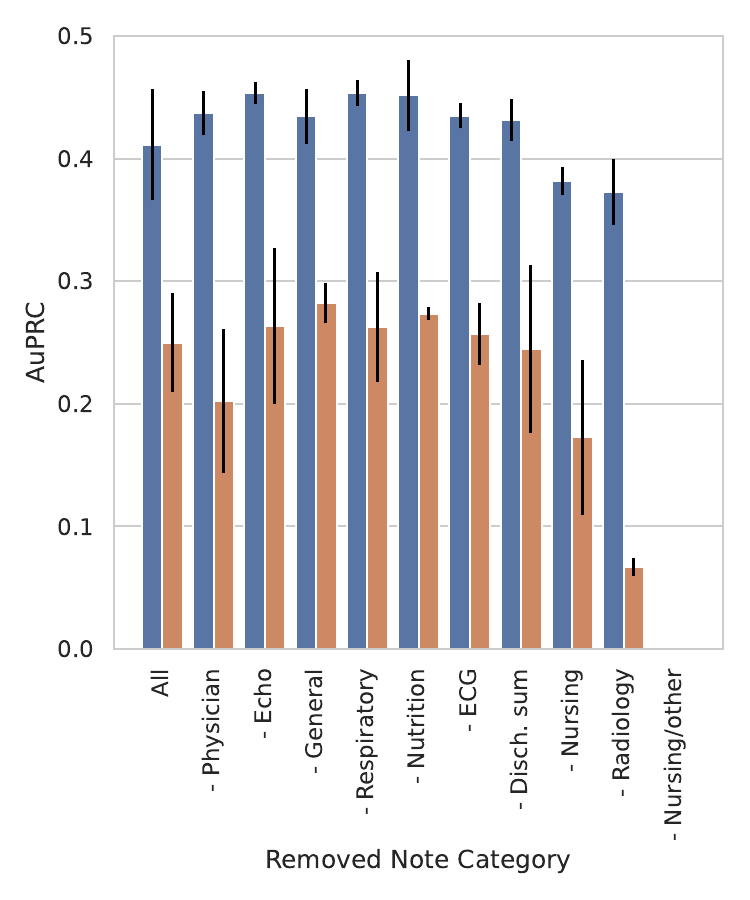}
        \caption{Mortality}
        \label{subfig:note_abl_mort}
    \end{subfigure}
    \begin{subfigure}[b]{0.49\textwidth}
        \centering
        \includegraphics[width=\textwidth]{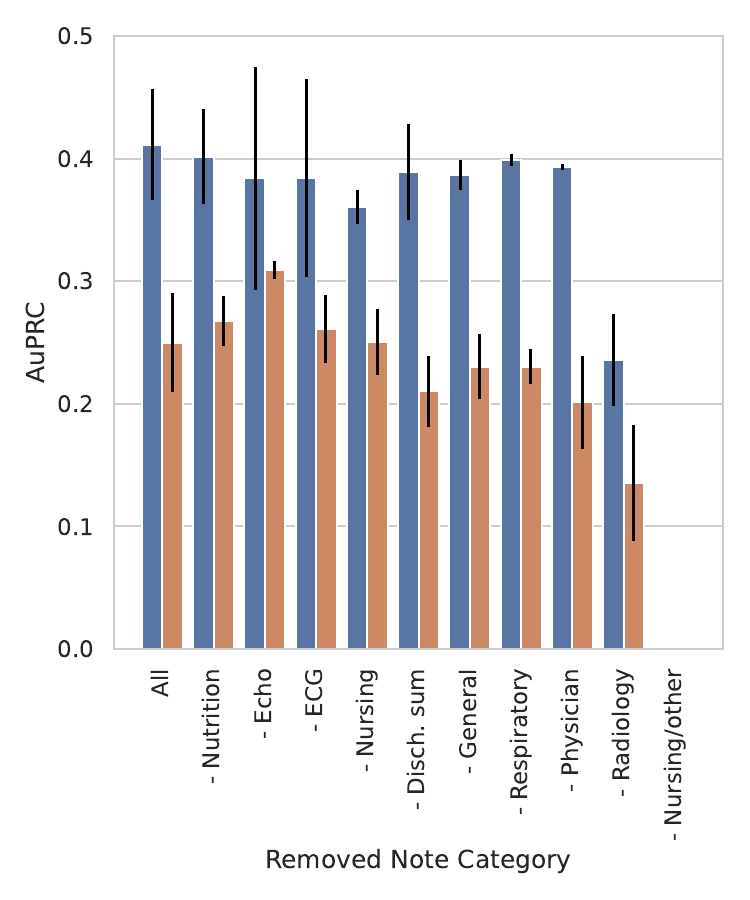}
        \caption{Decompensation}
        \label{subfig:note_abl_decomp}
    \end{subfigure}
\caption{Zero-shot AuPRC for mortality (blue) and decompensation (orange) for different sets of note types for \emph{MM-NCL}. We greedily remove note types from the left to the right based on mortality~(Fig.~\ref{subfig:note_abl_mort}) and decompensation~(Fig.~\ref{subfig:note_abl_decomp}) AuPRC. Removing the last remaining category (\texttt{Nursing/other} in both cases) leaves no training data for the text modality, so there is no result in the rightmost columns.}
    \label{fig:note-type-ablation}
\end{figure}

\paragraph{Ablation on Note Types} For each task, we optimized the set of note types used during training by greedily removing note types from the training data. In each run, we early-stopped based on the validation AuPRC of the task we optimized for. In Figure~\ref{fig:note-type-ablation}, we observe a strong effect of note type selection on model performance. Notable differences between the tasks are that \texttt{Physician} notes seem to be a lot more important for decompensation and that \texttt{Nursing} notes seem to be more important for mortality. To be expected was that the last remaining categories are \texttt{Radiology} and \texttt{Nursing/other} notes, as they make up 65\% of all considered notes. Also expected was that discharge summaries are more helpful for mortality as they highlight information relevant over the entire patient stay and tend to mention patient outcomes. Note that differences in the two plots for pre-training on \texttt{Nursing/other} only stem from optimizing the selection (including early stopping of the pre-training) for different tasks.

\section{Conclusion}

We proposed a new multi-modal contrastive loss function for clinical notes and time-series. Leveraging a soft neighborhood function we can train a multi-modal shared latent space, which exhibits strong performance under linear probing and facilitates unseen zero-shot classification performance in this application domain. Further research remains to validate our findings on other datasets and experiment with the inclusion of additional data modalities.

\subsubsection*{Acknowledgments}
We would like to thank Hugo Yèche for insightful discussions and valuable advice. This project was supported by grant \#2022-278 of the Strategic Focus Area “Personalized Health and Related Technologies (PHRT)” of the ETH Domain (Swiss Federal Institutes of Technology; to G.R.).

\newpage
\bibliography{iclr2024_conference}
\bibliographystyle{iclr2024_conference}

\clearpage
\appendix

\section{Data}
\label{app:data}
\subsection{Tasks}

We benchmark two mortality-related tasks defined by~\citet{Harutyunyan2019} on the MIMIC-III dataset~\citep{Johnson2016}. \citet{Harutyunyan2019} provide cohort selection and patient splits for training. Mortality information is directly extracted by~\citet{Harutyunyan2019} from the MIMIC-III patient metadata.

\textbf{In-Hospital Mortality} Given an ICU stay and its time-series $X^S$ the binary classification task is to predict, whether the patient died in the ICU or was discharged alive based on the first 48 hours of stay ($X^S[0:48]$).

\textbf{Decompensation} Given an ICU stay the \emph{online} binary classification task is to predict the onset of death at every hour of the patient's stay until death or discharged alive.

\subsection{Time-Series}

We prepare time-series data from MIMIC-III~\citep{Johnson2016} as published by~\citet{Harutyunyan2019}. Missing values are forwarded imputed if prior measurements are available. Data is standard-scaled and the remaining missing values after forward imputation are zero-imputed, which corresponds to a population mean imputation.

\subsection{Clinical Notes}
\label{app:clinical-notes}

We consider the same set of notes as \citet{jain2023knowledge} and provide their description and details on the clinical notes published in the MIMIC-III~\citep{Johnson2016} dataset in the \texttt{NOTEEVENTS} table.

In total, there are about 2 million individual text notes of 10 categories (\texttt{Discharge summary}, \texttt{ECG}, \texttt{Echo}, \texttt{General}, \texttt{Nursing}, \texttt{Nursing/other}, \texttt{Nutrition}, \texttt{Physician}, \texttt{Radiology} and \texttt{Respiratory}). The median length of such a note is 1090 characters and we observe a median of about 14 individual notes per patient (with 7 at the first quartile and 30 at the third quartile).

Each note is associated with a specific timestamp (\texttt{CHARTDATE} and \texttt{CHARTTIME}) during a single admission (\texttt{HADM\_ID}) of a given patient (\texttt{SUBJECT\_ID}). However, for a given patient admission we do not observe a note at every single time point on our resampled grid used during training. Some time points might have no clinical note associated with them, whereas others might have multiple, and they thus build an irregularly sampled time series of textual descriptions of the patient state.
 
\section{Training Details}
\label{app:training}
\subsection{Hyperparameters}

\begin{table*}[ht]
    \centering
    \caption{Hyperparameter ranges, chosen ones are in bold}
    \begin{tabular}{ll}
    \toprule
    \textbf{Parameter Name} & \textbf{Values} \\
    \midrule
    GRU hidden dimension                     & {[}128, \textbf{256}{]}\\
    GRU depth                                & {[}1,\textbf{2},3{]}                \\
    GRU dropout                              & {[}\textbf{0.1}, 0.2, 0.3{]}        \\
    Text Encoder number of hidden dimensions & {[}\textbf{1}{]}                    \\
    Text Encoder MLP hidden dimension              & {[}\textbf{4096}{]}                 \\
    Loss parameter $\alpha$       & {[}0.1, \textbf{0.3}, 0.5, 0.8, 0.95, 1.0{]}             \\
    Loss parameter $\beta$                 & {[}\textbf{2}, 4, 8, 16, 24, 48, 96{]} \\
    Window size                            & {[} 8, \textbf{16}, 24, 48 {]} \\
    Batch size                             &  \textbf{512} Patient Stays \\
    Learning Rate Adam~\citep{kingma2017adam} & $5e^{-4}$ \\
    \bottomrule
    \end{tabular}
    \label{tab:hyperparameters}
\end{table*}

We tuned the hyperparameters on the validation performances in several grid searches, refining iteratively over time. Table~\ref{tab:hyperparameters} shows an overview. We use PyTorch~\citep{paszke2019pytorch} for training the models. We trained all models with a single \texttt{NVIDIA RTX2080Ti} and an Intel \texttt{Xeon E5-2630v4} CPU.

For the target time selection hyperparameters we chose $b = 3$ and $a = 10$ for discharge summaries, $a = 30$ for radiology notes, and $a = 3$ for everything else. Those are set in time-steps, which translates to hours on the MIMIC-III Benchmark~\citep{Harutyunyan2019}.

\subsection{Model Selection}

We pretrain the model with \emph{MM-NCL} for 30 epochs, which has been tuned for best aggregated zero-shot task performance on the validation set. Future work should look into incorporating more efficient early-stopping methods such as \emph{LiDAR}~\citep{thilak2023lidar}, which enables early stopping online in a more efficient way.
 
\section{Evaluation}
\textbf{Ablation on window sizes}
\Cref{tab:winsz_ablation} shows the zero-shot performance of our model with different window sizes $w$.
\begin{table*}[tb]
\small
    \centering
    \caption{Zero-shot performance for different window sizes $w$ on the MIMIC-III Benchmark tasks. All values are in \% and denoted as $mean \pm std$. All hyperparameters except for window size are kept fixed.}
    \vspace{0.1cm}
    \begin{tabular}{l l l l l}
    \toprule
    Window Size & \multicolumn{2}{c}{\textbf{Mortality}} & \multicolumn{2}{c}{\textbf{Decompensation}}\\
    \cmidrule(lr){2-3} \cmidrule(lr){4-5}
    \textit{hours} & \textit{AuPRC} & \textit{AuROC} & \textit{AuPRC} & \textit{AuROC} \\

\midrule
	8  & 42.3 $\pm$ 1.3 & 80.6 $\pm$ 1.1 & 27.2 $\pm$ 0.5 & 87.1 $\pm$ 1.0 \\
	16 & 45.0 $\pm$ 3.2 & 81.2 $\pm$ 0.7 & 29.5 $\pm$ 0.5 & 87.5 $\pm$ 0.8 \\
	24 & 43.6 $\pm$ 2.8 & 79.0 $\pm$ 2.7 & 28.8 $\pm$ 2.1 & 88.1 $\pm$ 0.6 \\
	48 & 42.3 $\pm$ 0.2 & 80.1 $\pm$ 0.5 & 21.9 $\pm$ 3.9 & 86.8 $\pm$ 1.2 \\
    \bottomrule
    \end{tabular}
    \label{tab:winsz_ablation}
\end{table*}

\textbf{Model Prompts}
\label{app:model-prompts}

We present a collection of prompts used in the zero-shot classification prompt ensembles in Table~\ref{tab:prompts}. The prompts have been selected based on an inspection of the notes found in the MIMIC-III dataset~\citep{Johnson2016} conditioned on the class labels.

\begin{table*}[tb]
\small
    \centering
    \caption{Zero-shot performance for different window sizes $w$ on the MIMIC-III Benchmark tasks. All values are in \% and denoted as $mean \pm std$.}
    \vspace{0.1cm}
    \begin{tabular}{c | c | c}
    \toprule
    Class & \textbf{Mortality} & \textbf{Decompensation} \\
    \midrule
    \rotatebox[origin=c]{90}{ Positive Prompts } & \makecell{
        patient deceased
        \\ passed away
        \\ patient died
        \\ died
        \\ deceased
        \\ expired
        \\ condition: expired
        \\ care withdrawn} &
    \makecell{
        Discharge Condition: Expired     	
        \\ Expired                          	
        \\ died                             	
        \\ dnr   
    } \\
    \midrule
    \rotatebox[origin=c]{90}{ Negative Prompts } & \makecell{
        survived
        \\ stable
        \\ discharged
    } &
    \makecell{
        stable
        \\ stable condition
        \\ discharged today
    } \\
    \bottomrule
    \end{tabular}
    \label{tab:prompts}
\end{table*}

\end{document}